\pdfoutput=1

\documentclass[11pt]{article}

\usepackage[]{acl}
\usepackage{acl}
\usepackage{tikz}
\usetikzlibrary{shapes}
\usetikzlibrary{graphs,graphs.standard,quotes}

\usepackage{ifthen}
\usepackage{times}
\usepackage{latexsym}
\usepackage{graphicx}
\usepackage{amsmath}
\usepackage{amsfonts}
\usepackage{booktabs}
\usepackage{siunitx}
\usepackage{algorithm}
\usepackage{algorithmic}
\usepackage{hyperref}
\usepackage{url}
\usepackage{pifont}
\usepackage{booktabs}
\usepackage{marvosym}
\usepackage{tikz}
\usepackage{colortbl}
\usepackage{caption}
\usepackage{multirow}
\usepackage{subcaption}
\usepackage{inconsolata}
\usepackage{stmaryrd}
\usepackage[T1]{fontenc}

\usepackage[utf8]{inputenc}

\usepackage{microtype}

\definecolor{ourRed}{HTML}{E24A33}
\definecolor{ourBlue}{HTML}{348ABD}
\definecolor{ourPurple}{HTML}{988ED5}
\definecolor{ourGray}{HTML}{777777}
\definecolor{ourLightGray}{HTML}{B8B8B8}
\definecolor{ourYellow}{HTML}{FBC15E}
\definecolor{ourGreen}{HTML}{4D8951}
\definecolor{ourPink}{HTML}{FFB5B8}
\definecolor{oursteelblue}{HTML}{9BB8D7}
\definecolor{ourOrange}{HTML}{FDBA58}
\definecolor{ourWhite}{HTML}{FAFAFA}

\newcommand{\Figref}[1]{Figure~\ref{#1}}
\newcommand{\figref}[1]{Figure~\ref{#1}}
\newcommand{\Tabref}[1]{Table~\ref{#1}}
\newcommand{\tabref}[1]{Table~\ref{#1}}
\newcommand{\Secref}[1]{Section~\ref{#1}}
\newcommand{\secref}[1]{Section~\ref{#1}}
\newcommand{\Algref}[1]{Algorithm~\ref{#1}}
\newcommand{\Appref}[1]{Appendix~\ref{#1}}
\newcommand{\Eqnref}[1]{Eqn.~\ref{#1}}
\newcommand{\CE}[1]{\textsc{CE}}

\newcommand{\ourmethod}{distillation interchange intervention training objective}

\newcommand{\ourmethodOG}{interchange intervention training}

\newcommand{\ourmethodabbrOG}{IIT}
\newcommand{\ourmethodabbr}{\textsc{DIITO}}

\newcommand{\intinv}{\textsc{IntInv}}
\newcommand{\get}{\textsc{GetVals}}
\newcommand{\set}{\textsc{SetVals}}

\newcommand{\contribfootnote}[1]{%
  \begingroup
  \renewcommand{\thefootnote}{}\footnote{#1}%
  \addtocounter{footnote}{-1}%
  \endgroup
}
\title{Causal Distillation for Language Models}

\author{
  Zhengxuan Wu$^{\ast}{}^\mathparagraph$,
  Atticus Geiger$^{\ast}{}^\mathparagraph$,
  Joshua Rozner,
  Elisa Kreiss,
  Hanson Lu
  \AND
  Thomas Icard,
  Christopher Potts,
  Noah D.~Goodman
  \AND
  \\[-3ex] Stanford University \\
  \texttt{\{wuzhengx, atticusg\}@stanford.edu}\\
}

\begin{document}
\maketitle
\begin{abstract}
Distillation efforts have led to language models that are more compact and efficient without serious drops in performance. The standard approach to distillation trains a student model against two objectives: a task-specific objective (e.g., language modeling) and an imitation objective that encourages the hidden states of the student model to be similar to those of the larger teacher model. In this paper, we show that it is beneficial to augment distillation with a third objective that encourages the student to imitate the \emph{causal} dynamics of the teacher through a \emph{\ourmethod{}} (\ourmethodabbr{}). \ourmethodabbr\ pushes the student model to become a \emph{causal abstraction} of the teacher model -- a faithful model with simpler causal structure. \ourmethodabbr\ is fully differentiable, easily implemented, and combines flexibly with other objectives. Compared against standard distillation with the same setting, \ourmethodabbr\ results in lower perplexity on the WikiText-103M corpus (masked language modeling) and marked improvements on the GLUE benchmark (natural language understanding), SQuAD (question answering), and CoNLL-2003 (named entity recognition).\contribfootnote{$^{\ast}$Equal contribution. $^\mathparagraph$Correspondence authors.}\footnote{We release our code at~\url{https://github.com/frankaging/Causal-Distill}}
\end{abstract}

\section{Introduction}

Large pretrained language models have improved performance across a wide range of NLP tasks, but can be costly due to their large size. 
\emph{Distillation} seeks to reduce these costs while maintaining performance by training a simpler student model from a larger teacher model \cite{hinton2015distilling,sun-etal-2019-patient,sanh2019distilbert, jiao2019tinybert}. 

\citet{hinton2015distilling} propose model distillation with an objective that encourages the student to produce output logits similar to those of the teacher while also supervising with a task-specific objective (e.g., sequence classification).
\citet{sanh2019distilbert}, \citet{sun-etal-2019-patient}, and \citet{jiao2019tinybert} adapt this method, strengthening it with additional supervision to align internal representations between the two models. However, these approaches may push the student model to match all aspects of internal states of the teacher model irrespective of their \emph{causal} role in the network's computation. 
This motivates us to develop a method that focuses on aligning the \emph{causal} role of representations in the student and teacher models. 



We propose augmenting standard distillation with a new objective that pushes the student to become a \emph{causal abstraction} \citep{Beckers_Halpern_2019,beckers20a,geiger2021} of the teacher model: the simpler student will faithfully model the causal effect of teacher representations on output. 
To achieve this, we employ the \emph{\ourmethodOG} (\ourmethodabbrOG) method of \citet{geiger-etal-2021-iit}. 
The \emph{\ourmethod{}} (\ourmethodabbr{}) aligns a high-level student model with a low-level teacher model and performs \emph{interchange interventions} (swapping of aligned internal states); during training the high-level model is pushed to conform to the causal dynamics of the low-level model. 

\begin{figure*}[ht]
\centering
\resizebox{1.0\textwidth}{!}{
\input{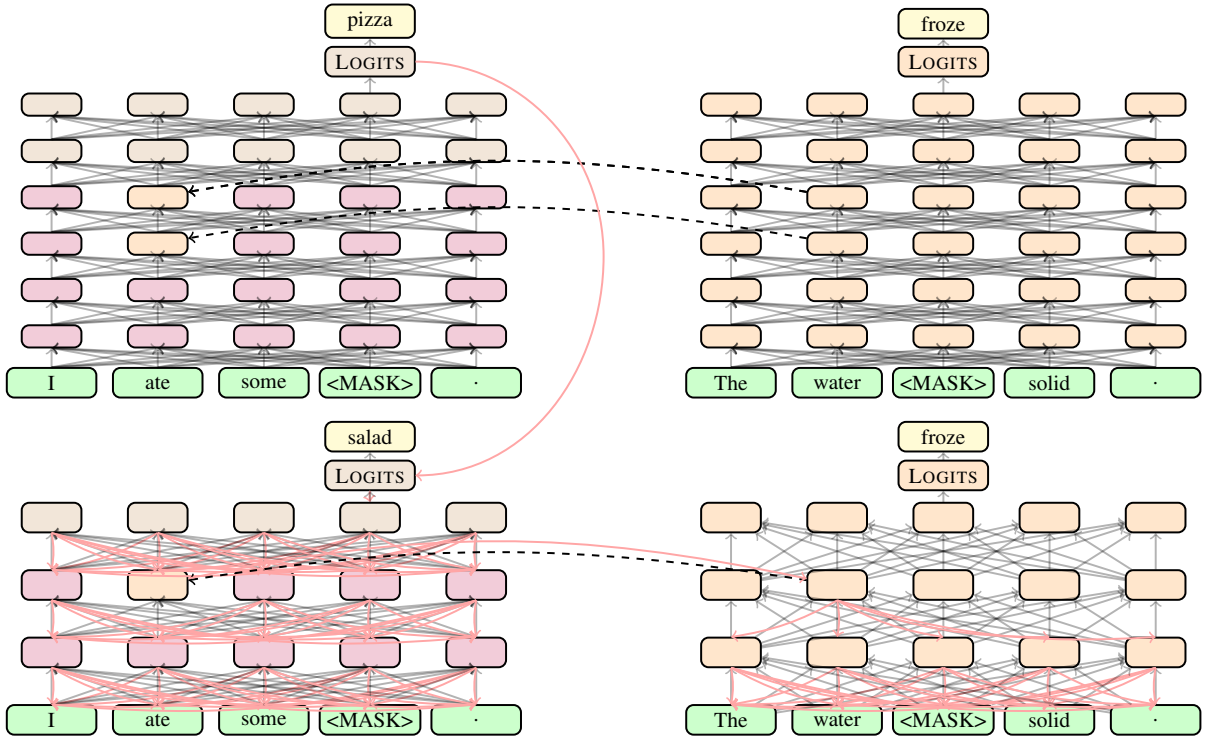}
}
\caption{An \ourmethodabbrOG{} update in the context of masked language modelling (MLM). The teacher network (top) has 6 layers and the student (bottom) has 3 layers, and we align layer 2 in the student with layers 3--4 in the teacher. Solid lines are feed-forward connections, red lines show the flow of backpropagation, and dashed lines indicate interchange interventions. In this case, the student originally predicted the token ``salad'' under the interchange intervention, while the teacher predicted the token ``pizza'' under an aligned interchange intervention. \ourmethodabbr{} trains the student to minimize the divergence between the student logits and the teacher logits under the interchange intervention. This updates the student to conform to causal dynamics of the teacher. }
\label{fig:mainfig:update}
\end{figure*}

\Figref{fig:mainfig:update} shows a schematic example of this process. Here, hidden layer 2 of the student model (bottom) is aligned with layers 3 and 4 of the teacher model. The figure depicts a single interchange intervention replacing aligned states in the left-hand models with those from the right-hand models. 
This results in a new network evolution that is shaped both by the original input and the interchanged hidden states. 
It can be interpreted as a certain kind of counterfactual as shown in \Figref{fig:mainfig:update}: what would the output be for the sentence ``I ate some ${\langle}\texttt{MASK}{\rangle}$.''\ if the activation values for the second token at the middle two layers were set to the values they have for the input ``The water ${\langle}\texttt{MASK}{\rangle}$ solid.''?
\ourmethodabbr\ then pushes the student model to output the same logits as the teacher, i.e., matching the teacher's output distribution under the counterfactual setup. 

To assess the contribution of distillation with \ourmethodabbr, we begin with $\text{BERT}_{\text{BASE}}$ \citep{devlin-etal-2019-bert} and distill it under various alignments between student and teacher while pretraining on the WikiText-103M corpus~\cite{merity2016pointer} achieving $-$2.24 perplexity on the MLM task compared to standard DistilBERT trained on the same data. We then fine-tune the best performing distilled models and find consistent performance improvements compared to standard DistilBERT trained with the same setting on the GLUE benchmark (+1.77\%), CoNLL-2003 name-entity recognition (+0.38\% on F1 score), and SQuAD v1.1 (+2.46\% on EM score).

\section{Related Work}

Distillation was first introduced in the context of computer vision~\citep{hinton2015distilling} and has since been widely explored for language models~\citep{sun-etal-2019-patient, sanh2019distilbert, jiao2019tinybert}. For example, \citet{sanh2019distilbert} propose to extract information not only from output probabilities of the last layer in the teacher model, but also from intermediate layers in the fine-tuning stage. Recently, \citet{rotman2021model} 
adapt causal analysis methods to estimate the effects of inputs on predictions to compress models for better domain adaptation. In contrast, we focus on imbuing the student with the causal structure of the teacher.

Interventions on neural networks were originally used as a structural analysis method aimed at illuminating neural representations and their role in network behavior \citep{CausalLM, pryzant-etal-2021-causal, vig2020causal, elazar-etal-2020, giulianelli-etal-2020-analysing,geiger-etal-2020-neural,geiger2021}. \citet{geiger-etal-2021-iit} extend these methods to network optimization. We contribute to this existing research by adapting intervention-based optimization to the task of language model distillation.

\section{Causal Distillation}\label{sec:counterfactuals}

Here, we define our distillation training procedure. See \Algref{alg:learning} for a summary.

\begin{algorithm}[t]
   \caption{\textbf{Causal Distillation via Interchange Intervention Training}}
\label{alg:learning}
\begin{algorithmic}
   \STATE {\bfseries Require:} 
   Student model $\mathcal{S}$, 
   teacher model $\mathcal{T}$, 
   student output neurons $\mathbf{N}^{\mathbf{y}}_{\mathcal{S}}$,
   alignment $\Pi$, 
   shuffled training dataset $\mathcal{D}$.
\STATE 1: $\mathcal{S}$.train()
\STATE 2: $\mathcal{T}$.eval()
\STATE 3: $\mathcal{D'}$ = random.shuffle($\mathcal{D}$)
\STATE 4: $\mathbf{N}^{\mathbf{y}}_{\mathcal{T}}$ = $\Pi(\mathbf{N}^{\mathbf{y}}_{\mathcal{S}})$
\STATE 5: \textbf{while} not converged \textbf{do}
\STATE 6: \ \ \textbf{for} $\{\mathbf{x}_{1}, \mathbf{y}_{1}\}$, $\{\mathbf{x}_{2}, \mathbf{y}_{2}\}$ \textbf{in} 
iter($\mathcal{D}$, $\mathcal{D'}$) \textbf{do}
\STATE 7: \ \ \ \ $\mathbf{N}_{\mathcal{S}} =$ sample\_student\_neurons()
\STATE 8: \ \ \ \ $\mathbf{N}_{\mathcal{T}}$ = $\Pi(\mathbf{N}_{\mathcal{S}})$
\STATE 9: \ \ \ \ \textbf{with} no\_grad:
\STATE 10: \ \ \ \ \ \ $\mathcal{T}_{a} = \set{}($
\STATE 11: \ \ \ \ \ \ \ \ \ \ $\mathcal{T}, \mathbf{N}_{\mathcal{T}}, \get{}(\mathcal{T}, \mathbf{x}_{1}, \mathbf{N}_{\mathcal{T}}))$
\STATE 12: \ \ \ \ \ \ $o_{\mathcal{T}} = \get{}(\mathcal{T}_{a}, \mathbf{x}_{2}, \mathbf{N}^{\mathbf{y}}_{\mathcal{T}})$
\STATE 13: \ \ $\mathcal{S}_{a} = \set{}($
\STATE 14: \ \ \ \ \ \ $\mathcal{S}, \mathbf{N}_{\mathcal{S}}, \get{}(\mathcal{S}, \mathbf{x}_{1}, \mathbf{N}_{\mathcal{S}}))$
\STATE 15: \ \ $o_{\mathcal{S}} = \get{}(\mathcal{S}_{a}, \mathbf{x}_{2}, \mathbf{N}^{\mathbf{y}}_{\mathcal{S}})$
\STATE 16: \ \ $\mathcal{L}^{\ourmethodabbr} = \textbf{get\_loss}(o_{\mathcal{T}}, o_{\mathcal{S}})$
\STATE 17: \ \ Calculate $\mathcal{L}_{\textmd{MLM}}$, $\mathcal{L}_{\textmd{CE}}$, $\mathcal{L}_{\textmd{Cos}}$ 
\STATE 18: \ \ $\mathcal{L}$ = $\mathcal{L}_{\textmd{MLM}}$ + $\mathcal{L}_{\textmd{CE}}$ + $\mathcal{L}_{\textmd{Cos}}$ + $\mathcal{L}^{\ourmethodabbr}$
\STATE 19: \ \ $\mathcal{L}$.backward()
\STATE 20: \ \ Step optimizer
\STATE 21: \textbf{end while}
\end{algorithmic}
\end{algorithm}

\textbf{\get{}}.
The \get{} operator is an activation-value retriever for a neural model. Given a neural model $\mathcal{M}$ containing a set of neurons $\mathbf{N}$ (an internal representations) and an appropriate input $\mathbf{x}$, $\get{}(\mathcal{M}, \mathbf{x}, \mathbf{N})$ is the set of values that $\mathbf{N}$ takes on when processing $\mathbf{x}$. In the case that $\mathbf{N}$ represents the neurons corresponding to the final output, $\get{}(\mathcal{M}, \mathbf{x}, \mathbf{N})$ is the output of model $\mathcal{M}$ when processing $\mathbf{x}$ (i.e., output from a standard forward call of a neural model). 

\textbf{$\set{}$}.
The $\set{}$ operator is a function generator that defines a new neural model with a computation graph that specifies an intervention on the original model $\mathcal{M}$ \citep{pearl,Spirtes2001}. $\set{}(\mathcal{M}, \mathbf{N}, \mathbf{v})$ is the new neural model where the neurons $\mathbf{N}$ are set to constant values $\mathbf{v}$. Because we overwrite neurons with $\mathbf{v}$ in-place, gradients can back-propagate through $\mathbf{v}$.


\textbf{Interchange Intervention}. An interchange intervention combines \get{} and $\set{}$ operations. First, we randomly sample a pair of examples from a training dataset $(\mathbf{x}_1, \mathbf{y}_1),(\mathbf{x}_2, \mathbf{y}_2) \in  \mathcal{D}$.
Next, where $\mathbf{N}$ is the set of neurons that we are targeting for intervention, we define $\mathcal{M}_{\mathbf{N}}^{\mathbf{x}_{1}}$ to abbreviate the new neural model as follows:
\begin{equation}\label{eq:causal-m}
\set{}
\big(
        \mathcal{M}, 
        \mathbf{N}, 
        \get{}(\mathcal{M}, \mathbf{x}_{1}, \mathbf{N})
\big)
\end{equation}
This is the version of $\mathcal{M}$ obtained from setting the values of $\mathbf{N}$ to be those we get from processing input $\mathbf{x}_{1}$. The interchange intervention targeting $\mathbf{N}$ with $\mathbf{x}_{1}$ as the source input and $\mathbf{x}_{2}$ as the base input
is then defined as follows:
\begin{multline}\label{eq:interventionloss-getval}
\intinv (\mathcal{M}, \mathbf{N}, \mathbf{x}_{1}, \mathbf{x}_{2}) 
\overset{\textnormal{def}}{=} \\
\get{}(
    \mathcal{M}_{\mathbf{N}}^{\mathbf{x}_{1}},    
    \mathbf{x}_{2},
    \mathbf{N}^{\mathbf{y}})
\end{multline}
where $\mathbf{N}^{\mathbf{y}}$ are the output neurons. In other words, $\intinv (\mathcal{M}, \mathbf{N}, \mathbf{x}_{1}, \mathbf{x}_{2})$ is the output state we get from $\mathcal{M}$ for input $\mathbf{x}_{2}$ but with the neurons $\mathbf{N}$ set to the values obtained when processing input $\mathbf{x}_{1}$.


\textbf{\ourmethodabbr}. 
\ourmethodabbr\ employs $\mathcal{T}$ as the teacher model, $\mathcal{S}$ as the student model, $\mathcal{D}$ as the training inputs to both models, and $\Pi$ as an alignment that maps sets of student neurons to sets of teacher neurons. For each set of student neurons $\mathbf{N}_{\mathcal{S}}$ in the domain of $\Pi$, we define \ourmethodabbr{} loss as:
\begin{multline}\label{eq:interventionloss}
\mathcal{L}^{\ourmethodabbr}_{\textup{CE}}
 \overset{\textnormal{def}}{=} \\
 \sum\limits_{\mathbf{x}_{1},\mathbf{x}_{2}\in \mathcal{D}} 
 \textsc{CE}_{\textsc{S}}
 \Big(
 \intinv(\mathcal{S}, \mathbf{N}_{\mathcal{S}}, \mathbf{x}_{1}, \mathbf{x}_{2}),
 \\[-3ex]
 \intinv(\mathcal{T}, \Pi(\mathbf{N}_{\mathcal{S}}),  \mathbf{x}_{1}, \mathbf{x}_{2})
 \Big)
\end{multline}
where $\textsc{CE}_{\textsc{S}}$ is the smoothed cross-entropy loss measuring the divergences of predictions, under interchange, between the teacher and the student model.

\textbf{Distillation Objectives}.
We adopt the standard distillation objectives from DistilBERT~\cite{sanh2019distilbert} (defined formally in \Appref{app:standard-objectives}): $\mathcal{L}_{\textmd{MLM}}$ for the task-specific loss for the student model, $\mathcal{L}_{\textmd{CE}}$ for the loss measuring the divergence between the student and teacher outputs on masked tokens, and $\mathcal{L}_{\textmd{Cos}}$ for the loss measuring the divergence between the student and teacher contextualized representations on masked tokens in the last layer. Our final training objective for the student is a linear combination of the four training objectives reviewed above:  $\mathcal{L}_{\textup{MLM}}$, $\mathcal{L}_{\textup{CE}}$, $\mathcal{L}_{\textmd{Cos}}$, and $\mathcal{L}^{\ourmethodabbr}_{\textup{CE}}$. In a further experiment, we introduce a fifth objective $\mathcal{L}^{\ourmethodabbr}_{\textmd{Cos}}$ which is identical to $\mathcal{L}_{\textmd{Cos}}$, except the teacher and student are undergoing interchange interventions (see \Appref{app:loss-cos-causal} for details).

\begin{table*}[htp]
\centering 
\resizebox{\linewidth}{!}{%
    \begin{tabular}{lccccccccc} \toprule
         &   & Pretraining &  WikiText &  GLUE & \multicolumn{2}{c}{CoNLL-2003} & \multicolumn{2}{c}{SQuAD v1.1}\\ 
              {Model}  & Layers & Tokens & Perplexity & Score & {acc} & {F1} & {EM} & {F1}\\
        \midrule
        $\text{BERT}_{\text{BASE}}$~\cite{devlin-etal-2019-bert} & 12 & 3.3B   & 10.27 (--)${}^{\dagger}$ & 82.75 (--) & 96.40 (--) & 92.40 (--) & 80.80 (--) & 88.50 (--)\\
         (Wikipedia+BookCorpus)& & &   &   &  & &  & \\
        {DistilBERT~\cite{sanh2019distilbert}} & 6 & 3.3B  & 17.48 (--)${}^{\dagger}$ & 79.59 (--) & 98.39 (--)${}^{\dagger}$ & 93.10 (--)${}^{\dagger}$ & 77.70 (--) & 85.80 (--) \\
         (Wikipedia+BookCorpus)& & &  &  &    & &  & \\
        \midrule
        \midrule
        {DistilBERT} (WikiText) & 3  & 0.1B & 29.51 (0.32) & 67.42 (1.10) & 97.88 (0.04) & 88.89 (0.29) & 26.04 (0.93) & 68.38 (0.77) \\
        $\ourmethodabbr_{\texttt{MIDDLE}}$ (WikiText) & 3 & 0.1B   & 26.04 (0.93) & 69.30 (1.08) &  {98.03} (0.04) & 89.69 (0.18) & 58.74 (0.69) & 70.23 (0.57) \\
        $\ourmethodabbr_{\texttt{LATE}}$ (WikiText)& 3 & 0.1B  & 25.97 (0.63) & 69.01 (1.69) &  {98.03} (0.03) &  {89.82} (0.18) &  {58.75} (0.49) & 70.21 (0.41) \\
        $\ourmethodabbr_{\texttt{FULL}}$ (WikiText)& 3 & 0.1B  & 24.85 (0.58) & 69.36 (0.87) & 98.02 (0.03) & 89.67 (0.16) & 58.72 (0.67) &  {70.50} (0.56) \\ \midrule
        {DistilBERT} (WikiText)& 6  & 0.1B & 15.69 (1.51) & 75.80 (0.42) & 98.48 (0.03) & 92.12 (0.23) & 70.23 (0.75) & 79.99 (0.55) \\
        $\ourmethodabbr_{\texttt{MIDDLE}}$ (WikiText) & 6  & 0.1B & 14.32 (0.12) & 76.71 (0.47) &  {98.56} (0.04) &  {92.47} (0.19) & 71.93 (0.31) & 81.32 (0.23) \\
        $\ourmethodabbr_{\texttt{LATE}}$ (WikiText)& 6 & 0.1B  & 14.93 (0.23) & 76.80 (0.34) & 98.51 (0.02) & 92.36 (0.27) & 71.47 (0.28) & 81.01 (0.23) \\
        $\ourmethodabbr_{\texttt{FULL}}$ (WikiText) & 6  & 0.1B & 13.59 (0.25) & 76.67 (0.21) & 98.53 (0.04) & 92.35 (0.24) &  {71.96} (0.29) &  {81.33} (0.25) \\ \midrule
        $\ourmethodabbr_{\texttt{FULL}}$+Random (WikiText)& 6  & 0.1B & 13.95 (0.18) & 76.84 (0.29) & 98.54 (0.03) & 92.41 (0.24) & 71.90 (0.54) & 81.27 (0.39) \\
        $\ourmethodabbr_{\texttt{FULL}}$+Masked (WikiText)& 6  & 0.1B & 13.99 (0.16) & 76.80 (0.32) &  {98.55} (0.03) &  {92.45} (0.18) & 71.77 (0.59) & 81.09 (0.42) \\
        $\ourmethodabbr_{\texttt{FULL}}$+$\mathcal{L}^{\ourmethodabbr}_{\textmd{Cos}}$ (WikiText) & 6  & 0.1B & 13.45 (0.19) & 77.14 (0.37) & 98.54 (0.04) & 92.35 (0.24) &  {71.94} (0.31) &  {81.35} (0.23)  \\ \bottomrule
    \end{tabular}
    }

  \caption{Performance on the development sets of the WikiText, GLUE benchmark, CoNLL-2003 corpus for the name-entity recognition task, and SQuAD v1.1 for the question answering task. The score is the averaged performance scores with standard deviation (SD) for all tasks across 15 distinct runs. ${}^{\dagger}$Numbers are imputed from released models on Hugging Face \citep{wolf-etal-2020-transformers}.
  }\label{tab:glue-ner-qa-results}
\end{table*}

\section{Experimental Set-up}

We adapt the open-source Hugging Face implementation for model distillation \citep{wolf-etal-2020-transformers}.\footnote{\url{https://github.com/huggingface/transformers}} We distill our models on the MLM pretraining task~\cite{devlin-etal-2019-bert}. We use large gradient accumulations over batches as in~\citet{sanh2019distilbert} for better performance. Specifically, we distill all models for three epochs for an effective batch size of 240. In contrast to the setting of 4K per batch in BERT~\citep{devlin-etal-2019-bert} and DistilBERT~\citep{sanh2019distilbert}, we found that small effective batch size works better for smaller dataset. 
We weight all objectives equally for all experiments. With our new objectives, the distillation takes approximately 9 hours on 4 NVIDIA A100 GPUs.

\textbf{Student and Teacher Models}. Our two students have the standard BERT architecture, with 12 heads with a hidden dimension of 768. The larger student has 6 layers, the smaller 3 layers. Our pretrained teacher has the same architecture, except with 12 layers. Following practices introduced by~\citet{sanh2019distilbert}, we initialize our student model with weights from skipped layers (one out of four layers) in the teacher model. We use WikiText for distillation to simulate a practical situation with a limited computation budget. We leave the exploration of our method on larger datasets for future research.

\textbf{Alignment}. Our teacher and student BERT models create columns of neural representations above each token with each row created by the feed-forward layer of a Transformer block, as in \figref{fig:mainfig:update}. We define $L_{\mathcal{T}}$ and $L_{\mathcal{S}}$ to be the number of layers in the student and teacher, respectively. In addition, we define $\mathcal{S}_i^j$ and $\mathcal{T}_i^j$ to be the representations in the $i$th row and $j$th column in the student and teacher, respectively. An alignment $\Pi$ is a partial function from student representations to sets of teacher representations. 
We test three alignments:
\begin{description}\setlength{\itemsep}{0pt}
\item[\texttt{FULL}] $\Pi$ is defined on all student representations: 
$\Pi(\mathcal{S}_i^j) = \{\mathcal{T}_{4i+k}^j: 0 \leq k < L_{\mathcal{T}}/L_{\mathcal{S}}\}$
\item[\texttt{MIDDLE}] $\Pi$ is defined for the row $L_{\mathcal{S}} \sslash 2$: $\Pi(\mathcal{S}_{L_{\mathcal{S}} \sslash 2}^{j}) = \{\mathcal{T}_{L_{\mathcal{T}} \sslash 2}^{j}\}$
\item[\texttt{LATE}]  $\Pi$ is defined on the student representations in the first and second rows: \\
$\Pi(\mathcal{S}^{j}_{1}) = \{\mathcal{T}^{j}_{L_{\mathcal{T}}-2 }\}$ and $\Pi(\mathcal{S}^{j}_{2}) = \{\mathcal{T}^{j}_{L_{\mathcal{T}}-1}\}$
\end{description}

For each training iteration, we randomly select one aligned student layer to perform the interchange intervention, and we randomly select 30\% of token embeddings for alignment for each sequence. We experiment with three conditions with the \textbf{\texttt{FULL}} alignment: consecutive tokens ($\textbf{\ourmethodabbr}_{\textbf{\texttt{FULL}}}$), random tokens ($\textbf{\ourmethodabbr}_{\textbf{\texttt{FULL}}}$+\textbf{Random}) and masked tokens ($\textbf{\ourmethodabbr}_{\textbf{\texttt{FULL}}}$+\textbf{Masked}). We also 
include $\mathcal{L}^{\ourmethodabbr}_{\textmd{Cos}}$ to the \textbf{\texttt{FULL}} alignment ($\textbf{\ourmethodabbr}_{\textbf{\texttt{FULL}}}$+$\mathcal{L}^{\textbf{\ourmethodabbr}}_{\textbf{\textmd{Cos}}}$).



\section{Results}

\textbf{Language Modeling}. We first evaluate our models using perplexity on the held-out evaluation data from WikiText. As shown in \Tabref{tab:glue-ner-qa-results}, \ourmethodabbr\ brings performance gains for all alignments. Our best result is from the \textbf{\texttt{FULL}} alignment with the $\mathcal{L}_{\textmd{Cos}}$ ($\textbf{\ourmethodabbr}_{\textbf{\texttt{FULL}}}$+$\mathcal{L}^{\textbf{\ourmethodabbr}}_{\textbf{\textmd{Cos}}}$), which has $-$2.24 perplexity compared to standard DistilBERT trained with the same amount of data.


\textbf{GLUE}. The GLUE benchmark~\cite{wang2018glue} covers different natural language understanding tasks. We report averaged GLUE scores on the development sets by fine-tuning our distilled models in \tabref{tab:glue-ner-qa-results}. Individual task performance scores for each GLUE task are included in \Tabref{tab:glue-results-low} in the Appendix. The results suggest that distilled models with \ourmethodabbr{} lead to consistent improvements over standard DistilBERT trained under the same setting, with our best result ($\textbf{\ourmethodabbr}_{\textbf{\texttt{FULL}}}$+$\mathcal{L}^{\textbf{\ourmethodabbr}}_{\textbf{\textmd{Cos}}}$) being +1.77\% higher.


\textbf{Named Entity Recognition}.
We also evaluate our models on the CoNLL-2003 Named Entity Recognition task \citep{tjong-kim-sang-de-meulder-2003-introduction}. We report accuracy and Macro-F1 scores on the development sets. We fine-tune our models for three epochs. Our best performing model ($\textbf{\ourmethodabbr}_{\textbf{\texttt{MIDDLE}}}$) numerically surpasses not only standard DistilBERT (+0.38\% on F1 score) trained under the same setting, but also its teacher, $\text{BERT}_{\text{BASE}}$ (+0.05\% on F1 score). 
Though these improvements are small, in this case distillation produces a smaller model with \textit{better performance}.

\textbf{Question Answering}.
Finally, we evaluate on a question answering task, SQuAD v1.1~\citep{rajpurkar-etal-2016-squad}. We report Exact Match and Macro-F1 on the development sets as our evaluation metrics. We fine-tune our models for two epochs. \ourmethodabbr\ again yields marked improvements (\tabref{tab:glue-ner-qa-results}). Our best result is from the vanilla \textbf{\texttt{FULL}} alignment ($\textbf{\ourmethodabbr}_{\textbf{\texttt{FULL}}}$), with +2.46\% on standard DistilBERT trained under the same setting.

\begin{figure}[tp]
  \centering
  \includegraphics[width=1.0\columnwidth]{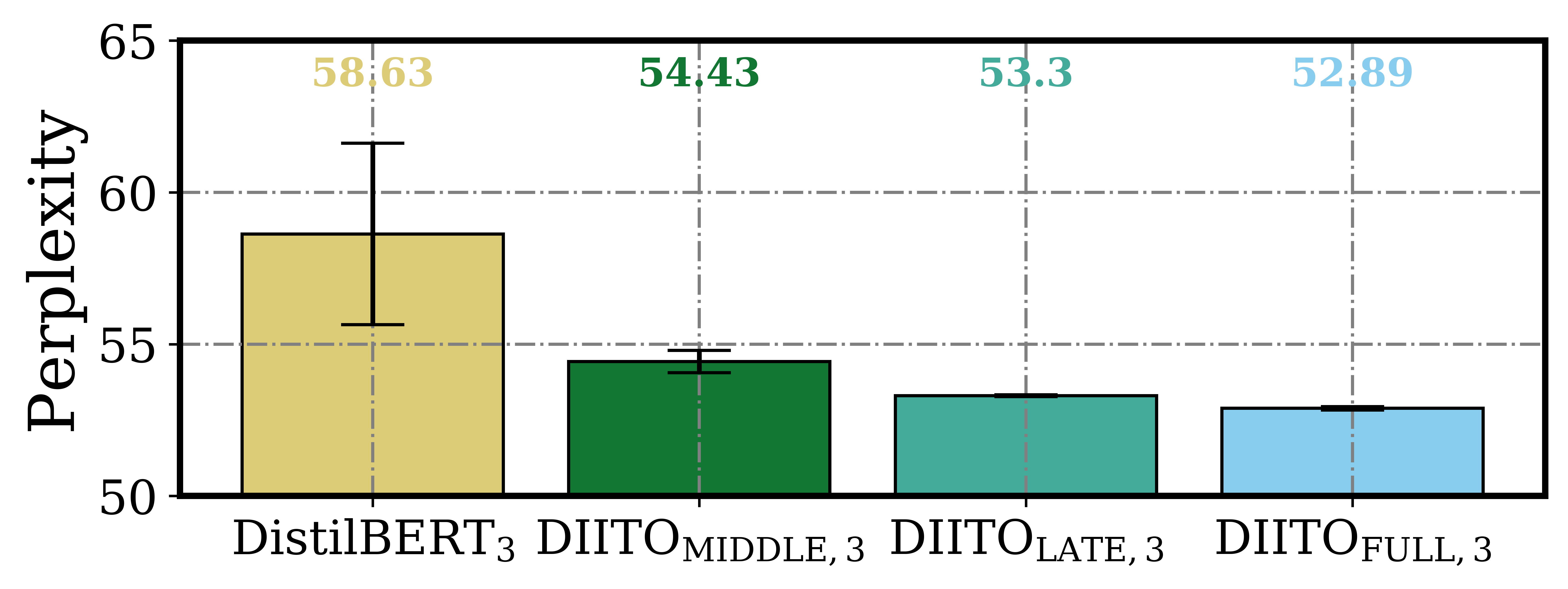}
  \caption{Perplexity score distribution for the development set of WikiText of models trained in a low-resource setting. The best model is the one with the richest alignment structure.}
  \label{fig:ppl_last_low}
\end{figure}

\textbf{Low-Resource Model Distillation}
We experiment with an extreme case in a low-resource setting where we only distill with 15\% of WikiText, keeping other experimental details constant. Our results suggest that \ourmethodabbr{} training is also beneficial in extremely low-resource settings (\Figref{fig:ppl_last_low}). 

\begin{figure}[tp]
  \centering
  \includegraphics[width=1.0\columnwidth]{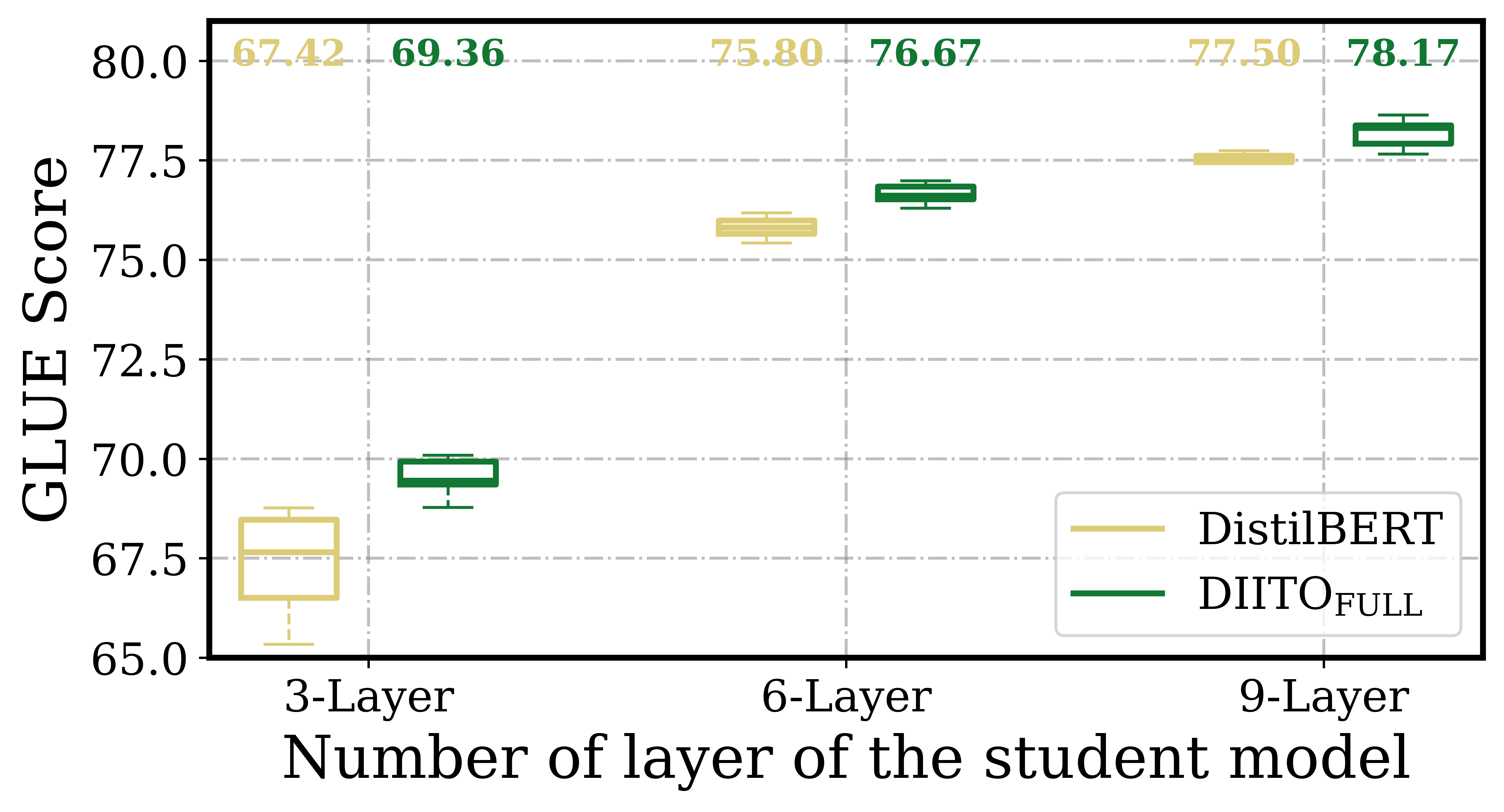}
  \caption{GLUE score distribution across 15 distinct runs of students in different sizes. Following the evaluation for BERT~\cite{devlin-etal-2019-bert}. we exclude WNLI for evaluation.}
  \label{fig:glue_layer}
\end{figure}

\textbf{Layer-wise Ablation}
We further study the effect of \ourmethodabbr{} training with respect to the size of the student model through a layer-wise ablation experiment. As shown in \Figref{fig:glue_layer}, we compare GLUE performance for models trained with standard distillation pipeline and with \ourmethodabbr{} training ($\textbf{\ourmethodabbr}_{\textbf{\texttt{FULL}}}$). Specifically, we compute the averaged GLUE scores following the same procedure described in \Secref{sec:reproduce}. Our results suggest that \ourmethodabbr{} training brings consistent improvements over GLUE tasks with smaller models booking the greatest gains.


\section{Conclusion}

In this paper, we explored distilling a teacher by training a student to capture the \emph{causal dynamics} of its computations.
Across a wide range of NLP tasks, we find that \ourmethodabbr\ leads to improvements, with the largest gains coming from the models that use the richest alignment between student and teacher. Our results also demonstrate that \ourmethodabbr\ performs on-par (maintaining 97\% of performance on GLUE tasks) with standard DistilBERT~\cite{sanh2019distilbert} while consuming 97\% less  training data. 
These findings suggest that \ourmethodabbr\ is a promising tool for effective model distillation.


\bibliographystyle{acl_natbib}
\bibliography{anthology,custom}

\appendix

\section{Appendix}
\label{sec:appendix}

\subsection{Standard Distillation Objectives} \label{app:standard-objectives}

In our setting, our teacher model $\mathcal{T}$ is a BERT model, and our student model $\mathcal{S}$ is a shallower BERT model with fewer layers. 

Assume that we randomly draw a training example $(\mathbf{x}_1, \mathbf{y}_1) \in \mathcal{D}$, where  $\mathbf{x}_{1}$ is the input to our models and $\mathbf{y}_{1}$ is the corresponding ground truth (the token prediction at each masked position). We denote the model predictions (output logits)  as $\mathcal{T}(\mathbf{x}_{1})$ and $\mathcal{S}(\mathbf{x}_{1})$. Additionally, we denote the contextualized representation for tokens for $\mathbf{x}_{1}$ at the last layer as $\text{BERT}_{\mathcal{T}}(\mathbf{x}_{1})$ and $\text{BERT}_{\mathcal{S}}(\mathbf{x}_{1})$.

We adopt the three standard distillation objectives of \citet{sanh2019distilbert}: 

\begin{description}\setlength{\itemsep}{0pt}
\item[$\mathcal{L}_{\textmd{MLM}}$] The masked language modeling loss of the student model calculated over all examples using the cross-entropy loss as follows:
\begin{equation}
\sum\limits_{\{\mathbf{x}_{1}, \mathbf{y}_{1}\} \in \mathcal{D}} \text{CE}(\mathcal{S}(
  \mathbf{x}_{1}), 
  \mathbf{y}_{1}
)
\end{equation}
\item[$\mathcal{L}_{\textmd{CE}}$] Following \citet{hinton2015distilling}, the smoothed cross-entropy loss measuring the divergence between the student and teacher outputs as follows:
\begin{equation}
\sum\limits_{\mathbf{x}_{1} \in \mathcal{D}} \textsc{CE}_{\textsc{S}}(\mathcal{S}(
  \mathbf{x}_{1}), 
  \mathcal{T}(\mathbf{x}_{1})
)
\end{equation}

\begin{table*}[ht]
\resizebox{\linewidth}{!}{%
  \centering 
    \newcommand{\spacer}{\hspace{24pt}}
    \begin{tabular}{lcccccccccc} \toprule
         &   & Pretraining & \multicolumn{8}{c}{General Language Understanding Evaluation (GLUE)} \\ 
              {Model}  & Layers & Tokens & {CoLA} & {MNLI} & {MRPC} & {QNLI} & {QQP} & {RTE} & {SST-2} & {STS-B}\\
        \midrule
        $\text{BERT}_{\text{BASE}}$~\cite{devlin-etal-2019-bert} & 12  & 3.3B   & 56.30 & 84.70 & 88.60 & 91.80 & 89.60 & 69.30 & 92.70 & 89.00 \\
         (Wikipedia+BookCorpus)& & &   &   &  & &  & \\
        {DistilBERT~\cite{sanh2019distilbert}} & 6 & 3.3B  & 51.30 & 82.10 & 87.50 & 89.20 & 88.50 & 59.90 & 91.30 & 86.90 \\
         (Wikipedia+BookCorpus)& & &   &   &  & &  & \\
        \midrule
        \midrule
        {DistilBERT} (WikiText) & 3  & 0.1B & 22.78 & 71.55 & 82.51 & 82.12 & 82.16 & 55.43 & 86.47 & 56.33 \\
        $\ourmethodabbr_{\texttt{MIDDLE}}$ (WikiText) & 3 & 0.1B   & 23.21 & 72.97 & 82.81 & 83.15 & 82.83 & 55.98 & 86.52 & 66.93 \\
        $\ourmethodabbr_{\texttt{LATE}}$ (WikiText) & 3 & 0.1B  & 24.12 & 72.80 & 82.16 & 82.88 & 82.85 & 57.29 & 87.31 & 62.65 \\
        $\ourmethodabbr_{\texttt{FULL}}$ (WikiText) & 3 & 0.1B  & 25.01 & 72.85 & 82.71 & 83.05 & 82.85 & 55.37 & 86.92 & 66.15 \\ \midrule
        {DistilBERT} (WikiText) & 6  & 0.1B & 40.43 & 78.95 & 87.45 & 84.76 & 84.96 & 60.10 & 89.38 & 80.40 \\
        $\ourmethodabbr_{\texttt{MIDDLE}}$ (WikiText) & 6  & 0.1B & 43.97 & 79.47 & 87.57 & 85.45 & 85.21 & 60.72 & 89.97 & 81.33 \\
        $\ourmethodabbr_{\texttt{LATE}}$ (WikiText) & 6 & 0.1B  & 43.93 & 79.49 & 87.70 & 85.79 & 85.22 & 60.14 & 90.31 & 81.79 \\
        $\ourmethodabbr_{\texttt{FULL}}$ (WikiText) & 6  & 0.1B & 43.43 & 79.66 & 88.17 & 85.57 & 85.28 & 59.95 & 90.01 & 81.26 \\ \midrule
        $\ourmethodabbr_{\texttt{FULL}}$+Random (WikiText) & 6  & 0.1B & 44.27 & 79.70 & 88.06 & 85.63 & 85.34 & 60.89 & 89.76 & 81.08 \\
        $\ourmethodabbr_{\texttt{FULL}}$+Masked (WikiText) & 6  & 0.1B & 43.39 & 79.63 & 87.88 & 85.61 & 85.30 & 61.06 & 89.97 & 81.58 \\
        $\ourmethodabbr_{\texttt{FULL}}$+$\mathcal{L}^{\ourmethodabbr}_{\textmd{Cos}}$ (WikiText) & 6 & 0.1B & 45.17 & 79.68 & 88.18 & 85.83 & 85.31 & 60.94 & 90.32 & 81.69 \\ \bottomrule
    \end{tabular}
    }
  \caption{Model performance results on the development sets of the GLUE benchmark. The GLUE score is the averaged performance scores across 15 distinct runs with precision aligned for a fair comparison. Following the evaluation for BERT~\cite{devlin-etal-2019-bert}, we exclude WNLI for evaluation.}
  \label{tab:glue-results-low}
\end{table*}

\item[$\mathcal{L}_{\textmd{Cos}}$] The cosine embedding loss defined in terms of the final hidden states of the teacher and the student as follows:
\begin{equation}
\sum\limits_{\mathbf{x}_{1} \in \mathcal{D}} \textsc{Cos}(
  \text{BERT}_{\mathcal{S}}(\mathbf{x}_{1}), 
  \text{BERT}_{\mathcal{T}}(\mathbf{x}_{1})
)
\end{equation}
\end{description}

As a result, comparing to standard DistilBERT, \ourmethodabbr\ essentially adds a new type of objective by pushing the student model to become a \emph{causal abstraction} of the teacher model.

\subsection{Causal Distillation Objectives} \label{app:loss-cos-causal}
In addition to our causal loss $\mathcal{L}^{\ourmethodabbr}_{\textup{CE}}$, we also propose a new loss $\mathcal{L}^{\ourmethodabbr}_{\textmd{Cos}}$ which is identical to $\mathcal{L}_{\textmd{Cos}}$ with interchange interventions. In this section, we provide a formal definition for $\mathcal{L}^{\ourmethodabbr}_{\textmd{Cos}}$. 

We denote our teacher and student models as $\mathcal{T}$ and $\mathcal{S}$ respectively. Using the notational conventions from \secref{sec:counterfactuals}, we use $\mathbf{N}^{\mathbf{y}}_{\mathcal{T}}$ and $\mathbf{N}^{\mathbf{y}}_{\mathcal{S}}$ to represent the neurons corresponding to the final output for each model. Likewise, we use $\mathbf{N}^{L_{\mathcal{T}}}_{\mathcal{T}}$ and $\mathbf{N}^{L_{\mathcal{S}}}_{\mathcal{S}}$ to represent the neurons representing contextualized representation for each token after the final BERT layer.

Assuming we randomly sample a pair of examples from a training dataset $(\mathbf{x}_1, \mathbf{y}_1),(\mathbf{x}_2, \mathbf{y}_2) \in  \mathcal{D}$, we can then rewrite our causal loss $\mathcal{L}^{\ourmethodabbr}_{\textup{CE}}$ by rearranging \Eqnref{eq:interventionloss-getval} and \Eqnref{eq:interventionloss} as follows: 
\begin{equation} \label{eq:causal-loss-connection}
\begin{aligned}
 \sum\limits_{\mathbf{x}_{1},\mathbf{x}_{2}\in \mathcal{D}} 
 \textsc{CE}_{\textsc{S}}
 \Big(
 & \get{}(
    \mathcal{M}_{\mathbf{S}}^{\mathbf{x}_{1}},    
    \mathbf{x}_{2},
    \mathbf{N}^{\mathbf{y}}_{\mathcal{S}}),
 \\[-3ex]
 & \get{}(
    \mathcal{M}_{\mathbf{T}}^{\mathbf{x}_{1}},    
    \mathbf{x}_{2},
    \mathbf{N}^{\mathbf{y}}_{\mathcal{T}})\Big)
\end{aligned}
\end{equation}
where $\mathcal{M}_{\mathbf{S}}^{\mathbf{x}_{i}}$ and $\mathcal{M}_{\mathbf{T}}^{\mathbf{x}_{i}}$ are derived as in \Eqnref{eq:causal-m} for each model respectively. Crucially, \Eqnref{eq:causal-loss-connection} can be regarded as the \emph{causal} form of the standard smoothed cross-entropy loss with interchange intervention. Likewise, we can further define the $\mathcal{L}^{\ourmethodabbr}_{\textmd{Cos}}$ as:
\begin{equation} \label{eq:causal-loss-connection-2}
\begin{aligned}
 \sum\limits_{\mathbf{x}_{1},\mathbf{x}_{2}\in \mathcal{D}} 
 \textsc{Cos}
 \Big(
 & \get{}(
    \mathcal{M}_{\mathbf{S}}^{\mathbf{x}_{1}},    
    \mathbf{x}_{2},
    \mathbf{N}^{L_{\mathcal{S}}}_{\mathcal{S}}),
 \\[-3ex]
 & \get{}(
    \mathcal{M}_{\mathbf{T}}^{\mathbf{x}_{1}},    
    \mathbf{x}_{2},
    \mathbf{N}^{L_{\mathcal{T}}}_{\mathcal{T}})\Big)
\end{aligned}
\end{equation}
with adjusted interchange alignments for $\mathbf{N}^{L_{\mathcal{T}}}_{\mathcal{T}}$ and $\mathbf{N}^{L_{\mathcal{S}}}_{\mathcal{S}}$. 

\subsection{Evaluation Set-up} 

\paragraph{GLUE} We fine-tune for 25 epochs for the smaller datasets (RTE and CoLA) and 3 epochs for the others. Following \citet{devlin-etal-2019-bert} and \citet{sanh2019distilbert}, we use Matthew's Correlation for CoLA, F1 for MRPC and QQP, Spearman correlation for STS-B, and accuracy for all the other tasks in GLUE. 

\subsection{Reproducibility} \label{sec:reproduce}

To foster reproduciblity and provide a fair comparison between methods, we distill BERT for each condition with three distinct random seeds. We then fine-tune each model with five distinct random seeds. Consequently, we report results aggregated from three distinct runs for the language modeling task, and 15 distinct runs for others.

\paragraph{Named Entity Recognition}
We follow the experimental set-up in the Hugging Face~\citep{wolf-etal-2020-transformers} repository for evaluation for the CoNLL-2003 Named Entity Recognition task \citep{tjong-kim-sang-de-meulder-2003-introduction}. For fine-tuning, we set the learning rate to $5e^{-5}$ with an effective batch size of 32 for three epochs.\footnote{For DistilBERT performance in~\Tabref{tab:glue-ner-qa-results} on CoNLL-2003, we evaluate with a publicly avaliable model downloaded from~\url{https://huggingface.co/delpart/distilbert-base-uncased-finetuned-ner}.}

\paragraph{Question Answering}
We use the experimental set-up of~\citet{sanh2019distilbert} for evaluation on SQuAD~v1.1~\citep{rajpurkar-etal-2016-squad}. For fine-tuning, we set the learning rate to $3e^{-5}$ with an effective batch size of 48 for two epochs. We set the stride to 128.

\end{document}